\documentclass{article}

\usepackage{microtype}
\usepackage{graphicx}
\usepackage{subcaption}
\usepackage{multirow}
\usepackage{booktabs} 

\usepackage{hyperref}

\usepackage{times}
\usepackage[preprint]{icml2026}

\usepackage{amsmath}
\usepackage{amssymb}
\usepackage{mathtools}
\usepackage{amsthm}
\usepackage{algorithm}
\usepackage{algorithmic}
\usepackage{listings}
\usepackage{xcolor}
\usepackage[capitalize,noabbrev]{cleveref}

\theoremstyle{plain}

\theoremstyle{definition}

\theoremstyle{remark}

\usepackage[textsize=tiny]{todonotes}

\icmltitlerunning{MIST-RL: Mutation-based Incremental Suite Testing via RL}

\begin{document}

\twocolumn[
  \icmltitle{MIST-RL: Mutation-based Incremental Suite Testing via \\Reinforcement Learning}



  \icmlsetsymbol{equal}{*}

\begin{icmlauthorlist}
    \icmlauthor{Sicheng Zhu}{fudan}
    \icmlauthor{Jiajun Wang}{ustc}
    \icmlauthor{Jiawei Ai}{xjtu}
    \icmlauthor{Xin Li}{scut}
  \end{icmlauthorlist}

  \icmlaffiliation{fudan}{Fudan University, Shanghai, China}
  \icmlaffiliation{ustc}{University of Science and Technology of China, Hefei, China}
  \icmlaffiliation{xjtu}{Xi'an Jiaotong University, Xi'an, China}
  \icmlaffiliation{scut}{South China University of Technology, Guangzhou, China}

  \icmlcorrespondingauthor{Sicheng Zhu}{24210240438@m.fudan.edu.cn}

  \icmlkeywords{Machine Learning, ICML, Software Testing, Reinforcement Learning, Large Language Models}

  \vskip 0.3in
]



\printAffiliationsAndNotice{}  

\begin{abstract}
Large Language Models (LLMs) often fail to generate correct code on the first attempt, which requires using generated unit tests as verifiers to validate the solutions. Despite the success of recent verification methods, they remain constrained by a ``scaling-by-quantity'' paradigm. This brute-force approach suffers from a critical limitation: it yields diminishing returns in fault detection while causing severe test redundancy. To address this, we propose \textbf{MIST-RL} (\textbf{M}utation-based \textbf{I}ncremental \textbf{S}uite \textbf{T}esting via \textbf{R}einforcement \textbf{L}earning), a framework that shifts the focus to ``scaling-by-utility''. We formulate test generation as a sequential decision process optimized via Group Relative Policy Optimization (GRPO). Specifically, we introduce a novel incremental mutation reward combined with dynamic penalties, which incentivizes the model to discover new faults while it suppresses functionally equivalent assertions. Experiments on HumanEval+ and MBPP+ demonstrate that MIST-RL outperforms state-of-the-art baselines. It achieves a \textbf{+28.5\%} higher mutation score while reducing the number of test cases by \textbf{19.3\%}. Furthermore, we show that these compact, high-utility tests serve as superior verifiers, which improves downstream code reranking accuracy on \textbf{HumanEval+} by \textbf{3.05\%} over the SOTA baseline with 10 candidate samples. The source code and data are provided in the supplementary material.
\end{abstract}

\section{Introduction}

Verifying the functional correctness of code generated by Large Language Models (LLMs) presents a core challenge in software engineering. While LLMs synthesize code effectively \citep{chen2021evaluating, austin2021program, roziere2023code, touvron2023llama}, they frequently suffer from hallucinations and subtle logic errors. Automated Unit Test Generation largely resolves this issue. These methods generate test cases to verify code functionality and serve as essential mechanisms for feedback, debugging, and candidate ranking (verifiers), which effectively bridges the gap between probabilistic generation and deterministic execution \citep{shi2022natural, li2022competition, cobbe2021training}.

Test-time scaling drives recent advancements in this domain. The prevailing assumption suggests that increasing the quantity of generated unit tests linearly translates to better fault detection and higher verification accuracy \citep{inala2022fault, huang2024enhancing}. Therefore, existing methods often resort to brute-force sampling where they generate massive \textbf{test suites} to cover potential execution paths.

\begin{figure*}[t]
    \centering
    \includegraphics[width=\linewidth]{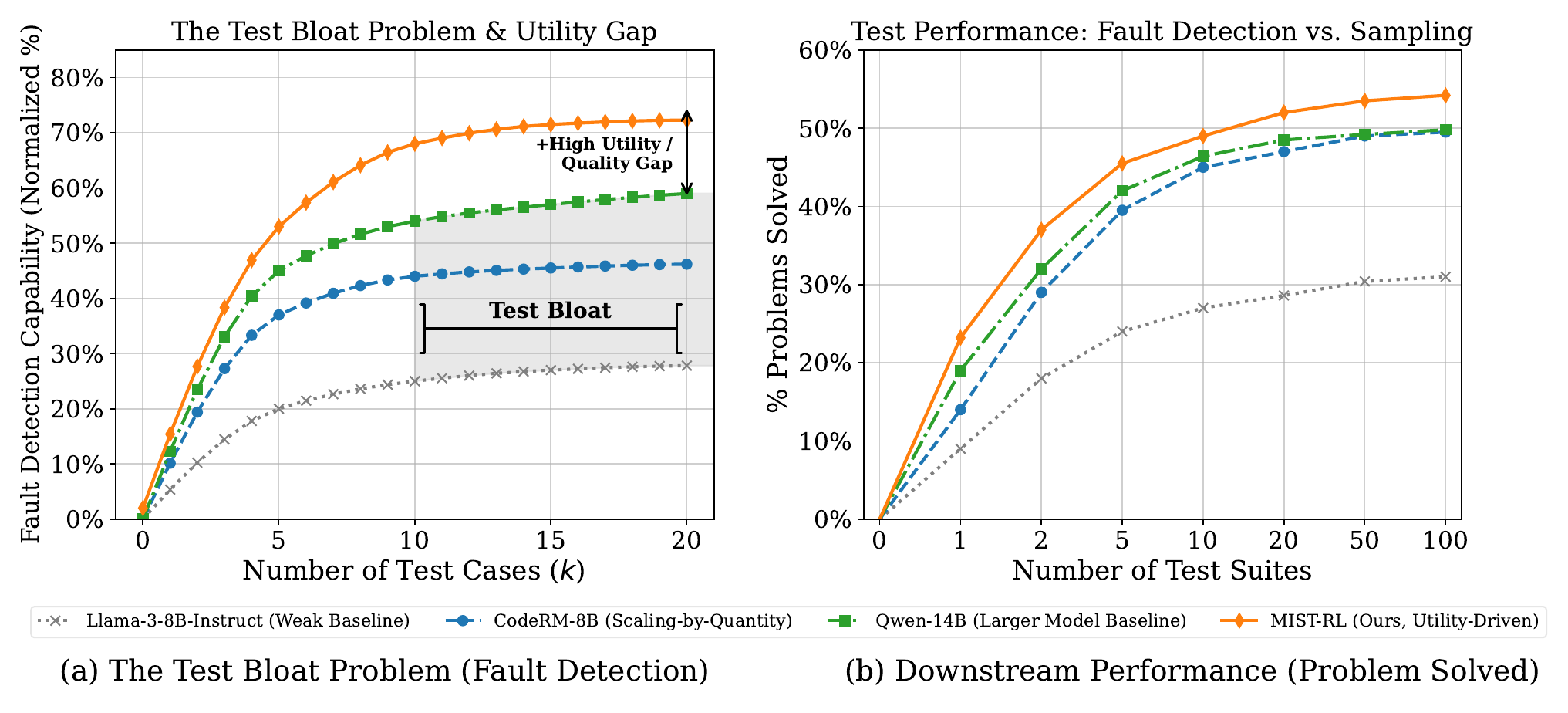}

    \caption{\textbf{Motivation: Quality Over Quantity.}
    \textbf{(a)} Existing ``scaling-by-quantity'' methods (Blue/Green lines), including the SOTA CodeRM-8B and the larger Qwen3-14B, exhibit rapid logarithmic saturation in Fault Detection Capability, indicating severe \textit{Semantic Redundancy} (shaded area). In contrast, \textbf{MIST-RL} (Orange) maintains a steep growth trajectory, creating a significant \textit{Utility Gap}.
    \textbf{(b)} This utility translates directly to downstream effectiveness. MIST-RL solves more problems (\% Problems Solved) with fewer \textbf{test suites} compared to baselines, validating that optimizing for marginal utility is more efficient than brute-force scaling.}
    \label{fig:motivation}
\end{figure*}

We challenge this ``\textbf{scaling-by-quantity}'' view. It faces a bottleneck: the law of diminishing returns caused by semantic redundancy. As \textbf{Figure~\ref{fig:motivation}(a)} shows, our empirical analysis reveals that simply increasing the number of test cases leads to saturation in fault detection capabilities. Even powerful models like Qwen3-14B \cite{yang2025qwen3} or specialized models like CodeRM-8B \cite{ma2025dynamic} eventually plateau. This indicates that the vast majority of later generated test cases are functionally identical—resulting in \textit{Test Bloat} \citep{yoo2012regression, singh2011analytical}. Such redundancy incurs unnecessary computational overhead and limits the effectiveness of the test suite as a verifier. \textbf{Figure~\ref{fig:motivation}(b)} confirms this, where baseline performance plateaus despite increased sampling.

The effectiveness of a test suite as a verifier hinges not on its size but on its \textbf{``aggressiveness''}—the ability to rigorously distinguish correct solutions from subtle bugs (e.g., off-by-one errors or boundary condition failures). A test suite can achieve high line coverage yet fail to reject plausible but incorrect code. The \textit{Mutation Score} best quantifies this discriminative power. We posit that the generation objective must explicitly prioritize test cases that kill hard-to-find mutants to improve downstream reranking, which maximizes the \textit{marginal utility} of verifying logic correctness.

To realize this utility-driven vision, we propose \textbf{MIST-RL}. This framework shifts the objective from ``scaling-by-quantity'' to ``\textbf{scaling-by-utility}''. We formulate test suite generation as a sequential decision process rather than a static completion task. By utilizing Reinforcement Learning (RL) with Group Relative Policy Optimization (GRPO) \citep{shao2024deepseekmath}, MIST-RL learns to construct compact yet aggressive test suites. The core of our approach is an \textit{incremental mutation reward} mechanism: the model receives incentives only when a new test case identifies faults (mutants) that survived all previous test cases. This dynamic feedback loop forces the policy to explore diverse failure modes and edge cases rather than repeating simple assertions.

We summarize the primary contributions of this paper below:

\begin{itemize}
    \item \textbf{Utility-Driven Generation:} We identify the limitations of prevailing quantity-oriented generation methods. We propose a new perspective that prioritizes the marginal fault-detection utility of individual \textbf{test cases} to combat test bloat.
    
    \item \textbf{RL-Based Incremental Framework:} We introduce MIST-RL, a reinforcement learning framework that integrates an incremental mutation reward system with dynamic redundancy penalties. This design aligns the generation policy with the goal of maximizing information gain per test case.
    
    \item \textbf{Enhanced Efficiency and Verification Quality:} Experiments on \textbf{HumanEval+ and MBPP+} demonstrate that MIST-RL achieves state-of-the-art mutation scores while significantly reducing test suite size (e.g., \textbf{-19.3\%} length reduction). We show that these compact, high-utility test suites serve as superior verifiers, which improves downstream code reranking accuracy by \textbf{3.05\%} over strong baselines.
\end{itemize}

\section{Background and Motivation}
\label{sec:background}

This section introduces the fundamental concepts of mutation testing and the prevailing ``scaling-by-quantity'' paradigm in LLM-based test generation. We then detail a pilot study that quantitatively reveals the inefficiencies of this paradigm—specifically the phenomenon of diminishing returns and test bloat—which directly motivates our proposed framework.

\subsection{Preliminary: Mutation Testing and Scaling Paradigms}

\paragraph{Mutation Testing.} 
Mutation testing~\cite{demillo2006hints, hamlet2006testing} evaluates the fault-detection capability of a test suite. It operates by injecting small synthetic faults (mutants $M$) into the source code $x$ using predefined mutation operators (e.g. changing an arithmetic operator $+$ to $-$). A test case $T$ is said to ``kill'' a mutant $m$ if the execution output of the mutated code differs from the original: $T(x) \neq T(m)$. The \textbf{Mutation Score (MS)} quantifies the effectiveness of a test suite $S$~\cite{jia2010analysis, papadakis2019mutation}:
\begin{equation}
    MS(S) = \frac{|\{m \in M \mid \exists T \in S, T \text{ kills } m\}|}{|M|}.
    \label{eq:ms}
\end{equation}

Unlike line coverage metrics, which merely quantify code reachability without guaranteeing that program states are correctly asserted~\cite{inozemtseva2014coverage}, mutation score provides a rigorous semantic assessment of whether the tests actually verify the code’s logic by simulating real logic errors~\cite{just2014mutants}. 
\textbf{Importantly}, unlike coverage-based metrics, mutation score directly measures unit tests' ability to distinguish correct behavior from subtle bugs, making it particularly suitable for verifier construction in code reranking.
In this work, we adopt MS as our primary optimization objective.

\paragraph{The ``Scaling-by-Quantity'' Paradigm.} 
Recent state-of-the-art (SOTA) approaches, most notably CodeRM~\cite{ma2025dynamic}, popularize the use of test-time scaling. These models are typically trained or prompted to generate a massive number of test cases to rerank candidate code solutions. The underlying assumption is that maximizing the volume of tests increases the probability of covering edge cases. While effective to a degree, this approach treats test generation as an independent, static prediction task, lacking an internal mechanism to evaluate the \textbf{marginal utility} of each new test case relative to the existing test suite.

\subsection{Motivation: The Diminishing Marginal Utility}
\label{subsec:motivation}

To investigate the efficiency of the scaling-by-quantity paradigm, we conducted a pilot study on the HumanEval+ dataset \citep{liu2023your} using the representative CodeRM-8B model. We prompted the model to generate a test suite of $N=20$ test cases for each problem and analyzed the relationship between test suite size and fault detection capability. 

As \textbf{Figure~\ref{fig:motivation}} shows, our analysis uncovers two critical bottlenecks that plague current methods:

\textbf{Observation 1: Saturation of Fault Detection.} 
Quantitative analysis reveals a strong \textbf{logarithmic saturation} behavior \citep{gao2023scaling}. As shown in Figure~\ref{fig:motivation}(a), the curve flattens rapidly. We find that the \textbf{top-25\%} of generated test cases (typically the first 5) contribute to over \textbf{85\%} of the total mutation score achieved by the full test suite. This heavy-tailed distribution indicates that the latter 75\% of the generation process is computationally wasteful and yields negligible gains in fault detection.

\textbf{Observation 2: Semantic Redundancy (Test Bloat).} 
Why does saturation occur so quickly? A closer inspection reveals severe semantic redundancy, also known as Test Bloat \citep{yoo2012regression}. Since the model generates test cases without explicit awareness of what has already been covered, it frequently produces assertions that are functionally identical (e.g., asserting $f(2)=4$ multiple times or testing the same logic branch with trivially different inputs). Our data shows that approximately \textbf{60\%} of assertions in the later stages target logic branches that were already fully killed by preceding test cases. This redundancy wastes inference compute and hampers downstream verification efficiency (Figure~\ref{fig:motivation}(b)).

\subsection{The Case for Incremental Reinforcement Learning}
\label{subsec:case_for_rl}

The observations above imply that a high-quality test suite should be \textbf{compact yet aggressive}. To achieve this, the generation process must be transformed from a static prediction task into a history-aware decision process. A new \textbf{test case} should only be generated if it targets a logic branch or potential bug that remains ``alive'' (uncovered).

To realize this, we require a generation policy explicitly driven by \textbf{marginal utility}. Instead of blindly sampling, the model should ideally operate under a dynamic feedback mechanism: receiving high incentives only when it successfully explores a \textit{new} failure mode, while incurring penalties for producing functionally equivalent assertions. Such a mechanism would naturally suppress redundancy and steer the model to identify complex boundary conditions that simple scaling fails to reach.

We propose \textbf{MIST-RL} to implement this vision. By reformulating test generation as an incremental Sequential Decision Process (SDP) and utilizing Reinforcement Learning, we effectively align the model's optimization objective with this utility-driven goal.

\section{Methodology}
\label{sec:methodology}

We present \textbf{MIST-RL}, a framework that transforms test generation from static text completion into a utility-driven sequential decision process. \textbf{Section 3.1} formulates the test generation task as a Markov Decision Process (MDP). \textbf{Section 3.2} details the construction of our mutation-based interaction environment. \textbf{Section 3.3} describes the core of our approach—the incremental reward mechanism designed to prioritize marginal fault-detection utility. Finally, \textbf{Section 3.4} describes the policy optimization using Group Relative Policy Optimization (GRPO).

\subsection{Problem Formulation}
\label{subsec:problem_formulation}
We model the generation of a \textbf{test suite} as a sequential decision process. Given a function under test (FUT) $x$, the policy $\pi_\theta$ iteratively generates a sequence of \textbf{test cases} $S = [T_1, T_2, \dots, T_K]$.

\textbf{Sequential Generation Process.} Unlike standard generation which produces the entire text at once, we view each test case $T_t$ as a distinct step in a trajectory. At each step $t$, the model conditions on the FUT $x$ and the previously generated test cases $T_{1:t-1}$ to generate the next test case $T_t$.

\textbf{History State.} To track progress, we define the \textbf{History State} $\mathcal{H}_t$ as the set of mutants killed by the partial sequence up to step $t$. Then:
\begin{equation}
    \mathcal{H}_t = \mathcal{H}_{t-1} \cup M(T_t), \quad \mathcal{H}_0 = \emptyset.
\end{equation}
Here, $M(T_t)$ denotes the set of mutants killed by the specific test case $T_t$. This state definition is crucial for calculating the incremental reward, as it allows us to distinguish between \textit{redundant} kills (mutants already in $\mathcal{H}_{t-1}$) and \textit{effective} kills (new mutants).

\subsection{State Space and Environment Construction}
\label{subsec:mutation_engine}

To enable high-throughput interaction and precise reward calculation during RL training, we developed a specialized, zero-dependency mutation engine based on Python's \textbf{Abstract Syntax Tree (AST)} module. Unlike heavyweight third-party tools \citep{lukasczyk2022pynguin, fraser2011evosuite}, our engine optimizes the fine-grained line mapping required for RL credit assignment.

\textbf{Standard Mutation Operators.} The engine implements six categories of operators to simulate diverse fault patterns, as detailed below \citep{just2014mutants, jia2010analysis}:
\begin{itemize}
    \item \textbf{AOR (Arithmetic)}: Perturbs numerical calculations (e.g., \texttt{+} $\to$ \texttt{-}).
    \item \textbf{ROR (Relational)}: Alters boundary conditions (e.g., \texttt{<} $\to$ \texttt{<=}), critical for testing edge cases.
    \item \textbf{LCR (Logical)}: Inverts control flow logic (e.g., \texttt{and} $\to$ \texttt{or}).
    \item \textbf{ASR (Assignment)}: Modifies state updates (e.g., \texttt{+=} $\to$ \texttt{-=}).
    \item \textbf{CRP (Constant)}: Perturbs literals (numbers, strings) and boolean flags.
    \item \textbf{UOI (Unary)}: Injects unary operators (e.g., flipping signs).
\end{itemize}

\textbf{Mitigating Equivalent Mutants.} To ensure that the generated mutants serve as valid learning targets, we enforce heuristic constraints during generation (e.g., verifying that a perturbed constant in CRP strictly differs from its original value). This filtering process minimizes the presence of \textit{equivalent mutants}—modifications that are semantically identical to the original code and thus unkillable. By eliminating these invalid targets, we ensure that the reward signal remains clean and actionable \citep{papadakis2019mutation}.

\textbf{Precise Line Mapping for Difficulty Weighting.} A critical challenge in RL-based testing is assigning appropriate importance to different mutants. To address this, we implement a mapping mechanism that aligns every generated mutant with its corresponding line number in the original source code. This allows us to assign difficulty weights $w_m$ (used in Eq.~\ref{eq:incremental_utility}) based on the code complexity of the mutated region, ensuring that the model receives higher utility rewards for covering complex logic branches.

\subsection{Incremental Reward Mechanism}
\label{subsec:reward}

Existing approaches often reward tests based on total coverage, leading to ``Test Bloat'' where models generate redundant tests for easy targets. To counter this, we propose a three-part reward system.

\begin{figure}[t]
    \centering
    \includegraphics[width=\columnwidth]{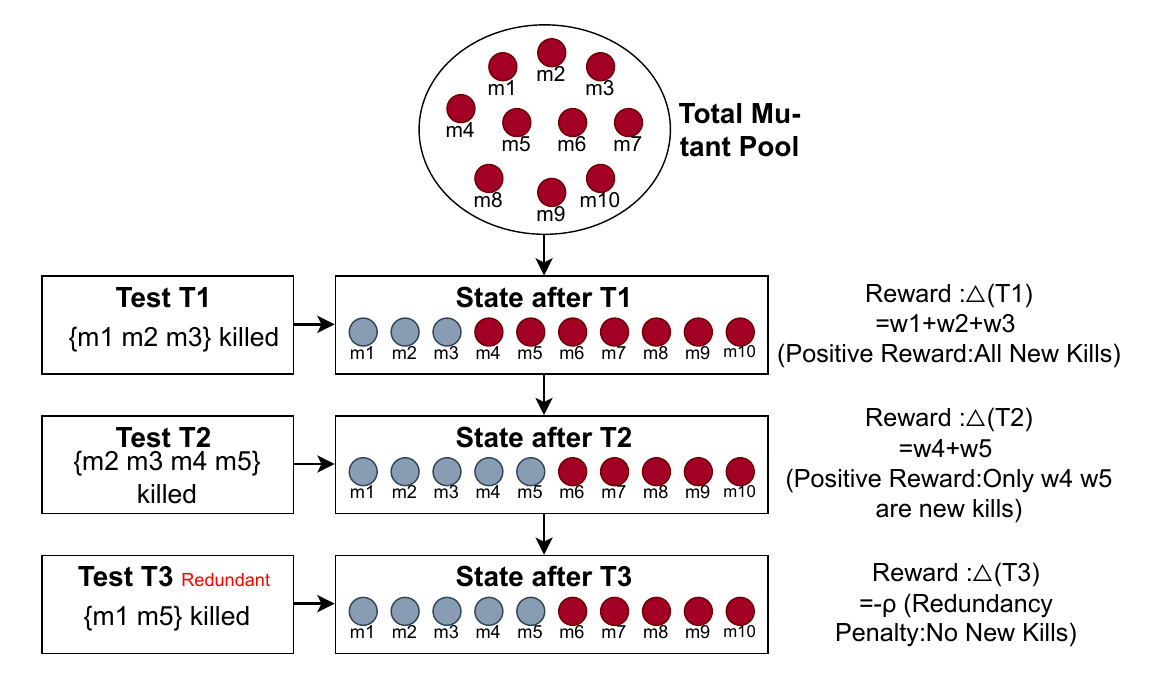}
    \caption{\textbf{Illustration of the Incremental Reward Mechanism.} The agent receives positive rewards only for killing \textit{new} mutants (grey circles). Redundant tests (e.g., T3) that fail to reduce the surviving mutant pool trigger a dynamic penalty ($-\rho$).}
    \label{fig:reward_mechanism}
\end{figure}

\subsubsection{Marginal Utility ($\Delta$)}
We define the reward of step $t$ strictly based on its \textbf{incremental contribution}. The marginal utility $\Delta(T_t \mid \mathcal{H}_{t-1})$ is the weighted sum of \textit{newly} killed mutants:
\begin{equation}
    \Delta(T_t \mid \mathcal{H}_{t-1}) = \sum_{m \in (M(T_t) \setminus \mathcal{H}_{t-1})} w_m.
\label{eq:incremental_utility}
\end{equation}
By using the set difference operator ($\setminus$), we ensure that a \textbf{test case} receives zero utility if it only kills mutants that were already eliminated by previous tests $T_{1 \dots t-1}$.

Theoretically, the mutation score function is \textbf{monotonically non-decreasing} and \textbf{submodular}, characterizing a property of diminishing returns. Consequently, the problem of selecting a \textbf{test suite} of size $K$ to maximize the mutation score is an NP-hard problem (essentially an instance of the \textbf{Maximum Coverage Problem}) \citep{yoo2012regression}. Our incremental reward mechanism encourages the policy to approximate a greedy algorithm for submodular maximization, enabling the model to efficiently identify the optimal subset of tests that covers the most diverse set of mutants.

\subsubsection{Dynamic Redundancy Penalty ($\rho_t$)}
To discourage the generation of infinite, low-value sequences, we introduce a step-dependent penalty $\rho_t$ that grows exponentially with the sequence length $t$:
\begin{equation}
    \rho_t = \rho_{base} \cdot e^{\gamma \cdot \frac{t}{K_{max}}},
\end{equation}
where $\gamma$ controls the growth rate. This forces the model to prioritize ``high-yield'' tests early in the sequence.

\subsubsection{Step-wise Piecewise Reward ($r_t$)}
Combining the above components, the final step reward $r_t$ is formulated as a piecewise function. To encourage the generation of semantically rich tests, we introduce a quality term $R_{qual}(T_t)$. This term employs a heuristic scoring mechanism based on the specificity and diversity of assertion primitives (e.g., prioritizing strict equality checks or exception handling over generic boolean assertions), with a saturation bound to prevent reward hacking via assertion stacking.

The final reward function is defined as:
\begin{equation}
    r_t = \begin{cases} 
    R_{fail}, & \text{if } \neg \mathbb{I}_{pass}(T_t); \\ 
    -\rho_t, & \text{if } \mathbb{I}_{pass}(T_t) \land (\Delta = 0); \\ 
    \alpha \cdot R_{qual}(T_t) + \beta \cdot \Delta, & \text{if } \mathbb{I}_{pass}(T_t) \land (\Delta > 0).
    \end{cases}
    \label{eq:piecewise_reward}
\end{equation}

The three scenarios are defined as follows:
\begin{itemize}
    \item \textbf{Case 1 (Failure):} If $T_t$ fails to compile or execute (e.g., syntax errors), a significant penalty $R_{fail}$ is applied, and the generation trajectory is truncated immediately.
    \item \textbf{Case 2 (Redundant):} If $T_t$ executes successfully but fails to kill any \textit{new} mutants ($\Delta=0$), it is deemed redundant. A dynamic penalty $-\rho_t$ is applied to suppress test bloat.
    \item \textbf{Case 3 (Effective):} If $T_t$ successfully contributes new information ($\Delta > 0$), the reward is a weighted sum of the code quality score $R_{qual}$ (ensuring assertion richness) and the marginal utility $\Delta$ (reflecting fault-detection difficulty).
\end{itemize}

\subsection{Optimization via GRPO}
\label{subsec:grpo}
We employ Group Relative Policy Optimization (GRPO) \citep{shao2024deepseekmath, guo2025deepseek} to optimize $\pi_\theta$. For each input $x$, we sample a group of $G$ outputs $\{O_1, \dots, O_G\}$. The total reward for a sequence is normalized by the square root of its effective length $K_{valid}$ to mitigate length bias:
\begin{equation}
    \mathcal{R}_{total}(S) = \frac{1}{\sqrt{K_{valid}}} \sum_{t=1}^{K_{valid}} r_t.
\end{equation}
The policy is updated by maximizing the advantage $A_i$, computed relative to the group mean:
\begin{equation}
    \small \mathcal{L}_{GRPO}(\theta) = -\frac{1}{G} \sum_{i=1}^{G} \frac{1}{|O_i|} \sum_{tok} \min \left( \frac{\pi_\theta}{\pi_{old}} A_i, \text{clip}(\dots) A_i \right),
\end{equation}
where $A_i = (\mathcal{R}_{total}(O_i) - \mu_{group}) / \sigma_{group}$. This reference-free approach eliminates the need for a separate value network (unlike PPO \citep{schulman2017proximal}), significantly reducing memory overhead during training.

\begin{table*}[t]
\centering
\caption{\textbf{Main Results on HumanEval+, MBPP+, and DS-1000.} MIST-RL consistently outperforms both the base model and strong baselines (including the larger Qwen3-14B) in fault detection (Mutant Kill Rate). It achieves these results while generating test suites that are significantly more concise than CodeRM-8B. The best results for \textbf{Source Pass} and \textbf{Mutant Kill} are highlighted in \textbf{bold}.}
\label{tab:main_results}

\vskip 0.1in 

\begin{small}
\begin{sc}
\resizebox{\textwidth}{!}{
\begin{tabular}{llccc}
\toprule
\textbf{Dataset} & \textbf{Model} & \textbf{Avg. Length} & \textbf{Source Pass (\%)}  & \textbf{Mutant Kill (\%)}  \\
\midrule
\multirow{4}{*}{\textbf{HumanEval+}} 
& Llama-3-8B-Instruct & 2.33 & 40.99 & 29.18 \\
& CodeRM-8B & 7.61 & 70.61 & 45.53 \\
& Qwen3-14B & 5.12 & 74.31 & 58.69 \\
& \textbf{MIST-RL (Ours)} & 6.14 & \textbf{74.86} & \textbf{74.03} \\
\midrule
\multirow{4}{*}{\textbf{MBPP+}} 
& Llama-3-8B-Instruct & 2.35 & 42.51 & 22.54 \\
& CodeRM-8B & 6.55 & 72.67 & 61.08 \\
& Qwen3-14B & 4.35 & \textbf{80.24} & 66.50 \\
& \textbf{MIST-RL (Ours)} & 5.17 & 77.54 & \textbf{70.27} \\
\midrule
\multirow{4}{*}{\textbf{DS-1000}} 
& Llama-3-8B-Instruct & 2.34 & 33.48 & 17.25 \\
& CodeRM-8B & 7.37 & 62.47 & 49.08 \\
& Qwen3-14B & 4.95 & 67.50 & 53.20 \\
& \textbf{MIST-RL (Ours)} & 5.78 & \textbf{68.12} & \textbf{57.90} \\
\bottomrule
\end{tabular}
}
\end{sc}
\end{small}
\end{table*}

\begin{figure*}[t]
    \centering
    \begin{subfigure}[b]{0.24\textwidth}
        \centering
        \includegraphics[width=\textwidth]{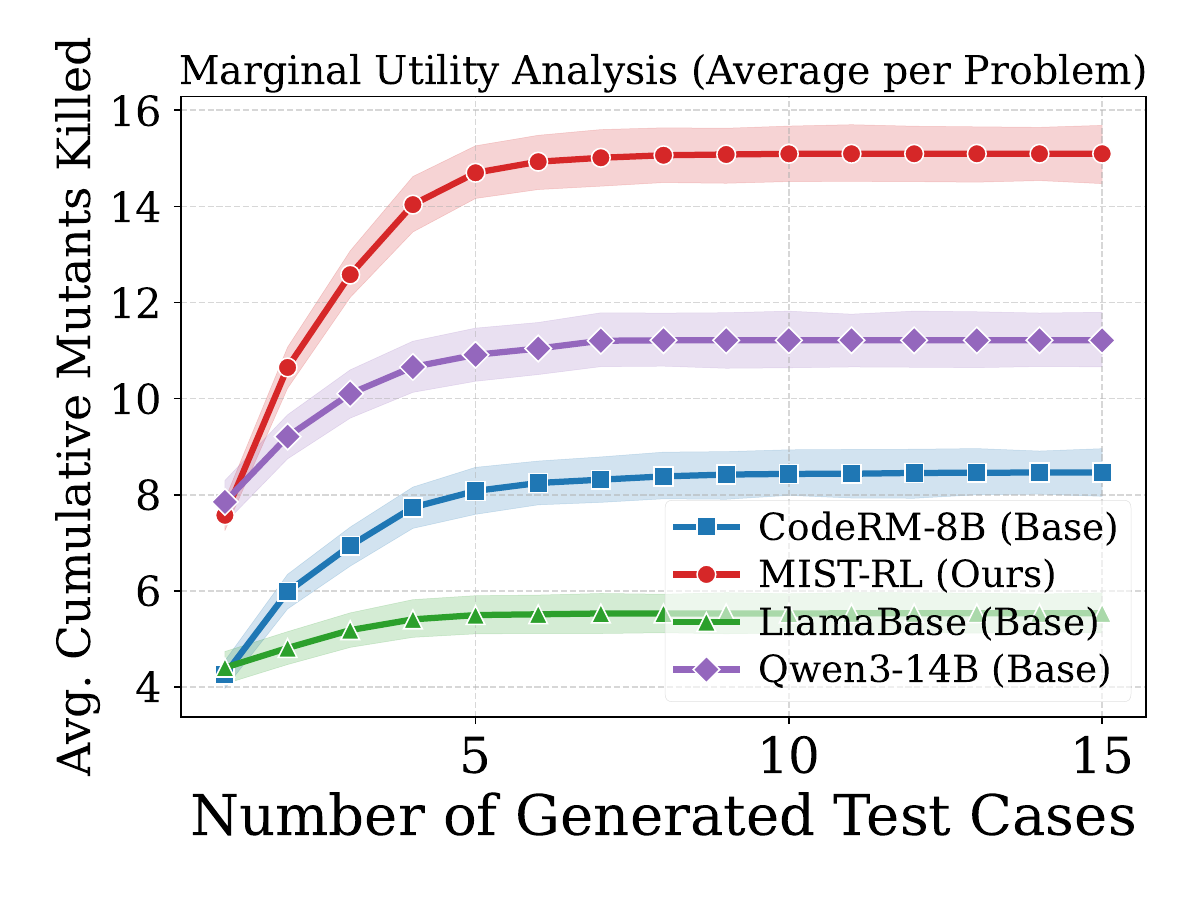}
        \caption{Avg. Utility (HE+)}
        \label{fig:utility_avg_he}
    \end{subfigure}
    \hfill
    \begin{subfigure}[b]{0.24\textwidth}
        \centering
        \includegraphics[width=\textwidth]{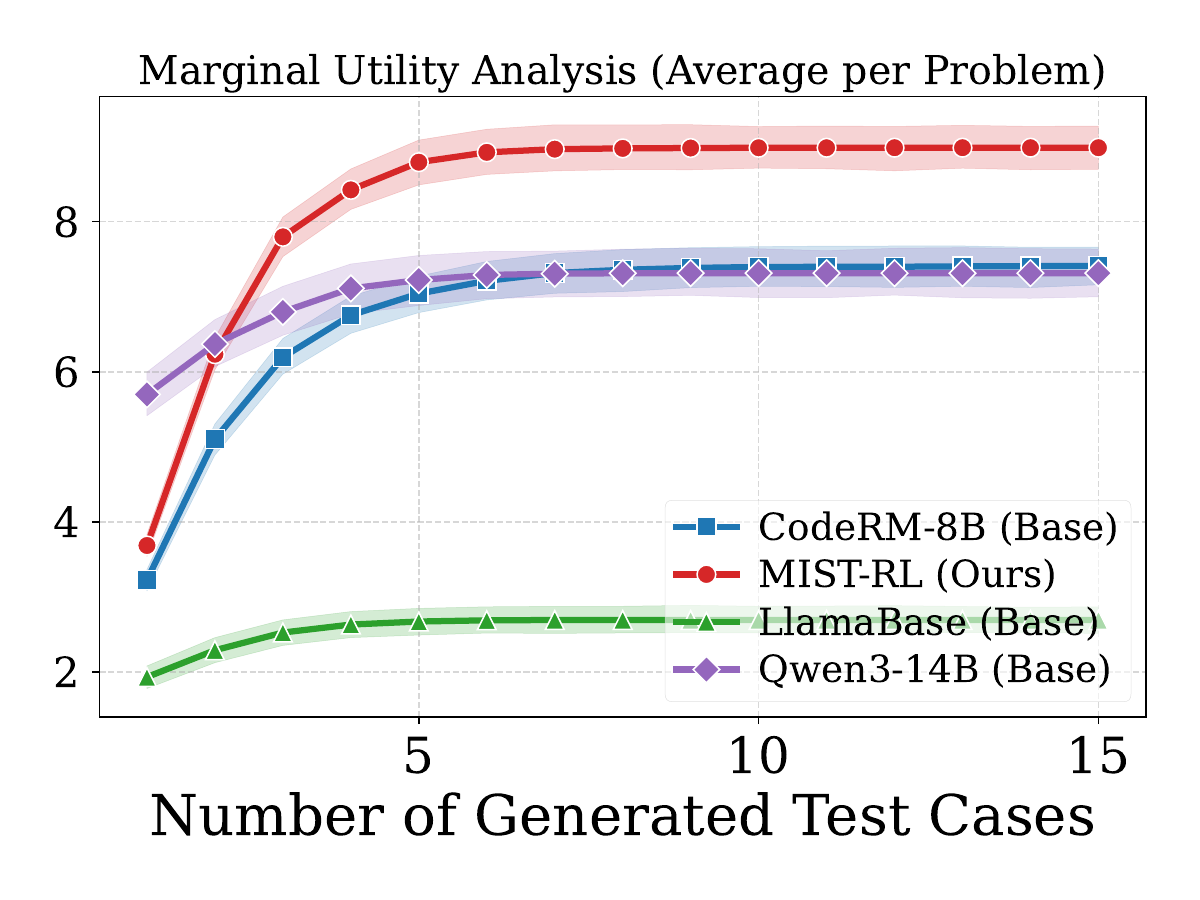}
        \caption{Avg. Utility (MBPP+)}
        \label{fig:utility_avg_mbpp}
    \end{subfigure}
    \hfill
    \begin{subfigure}[b]{0.24\textwidth}
        \centering
        \includegraphics[width=\textwidth]{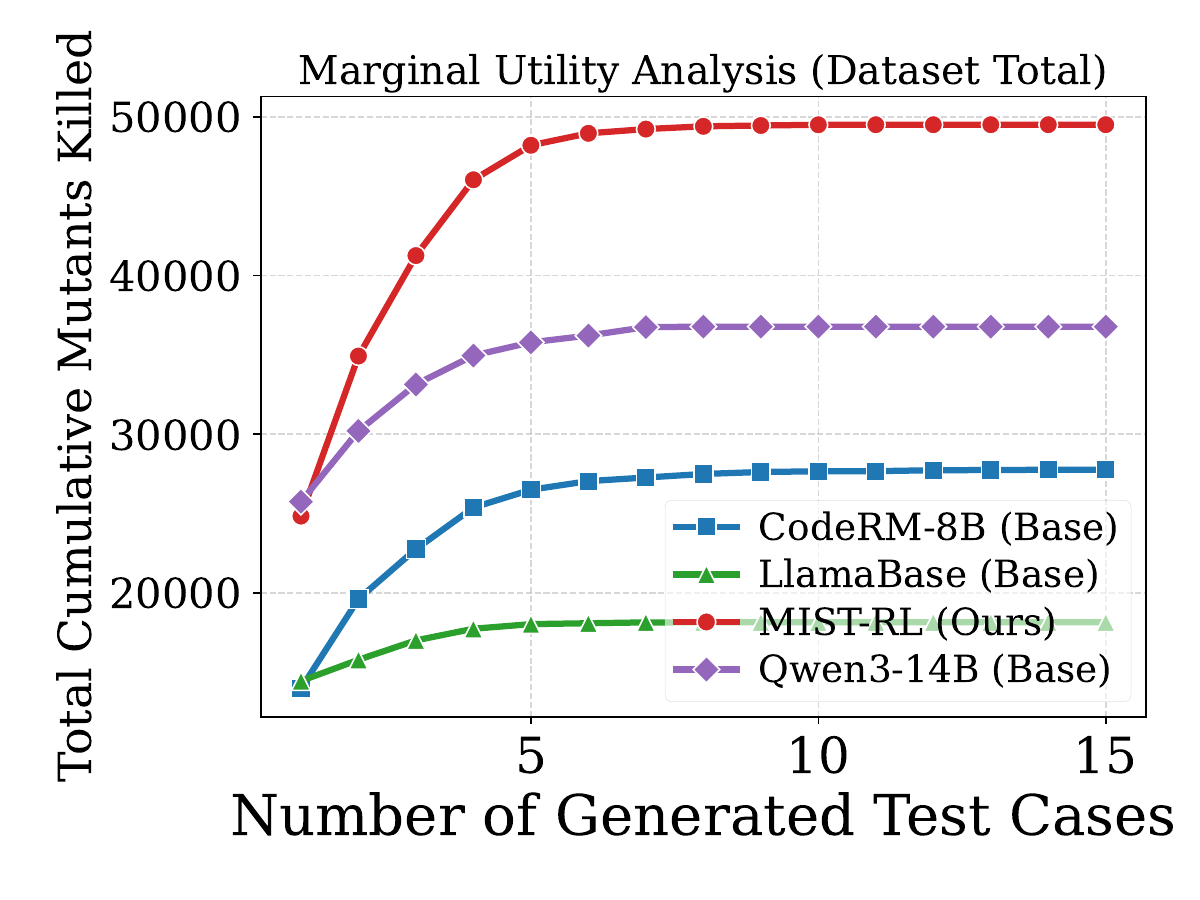}
        \caption{Total Utility (HE+)}
        \label{fig:utility_sum_he}
    \end{subfigure}
    \hfill
    \begin{subfigure}[b]{0.24\textwidth}
        \centering
        \includegraphics[width=\textwidth]{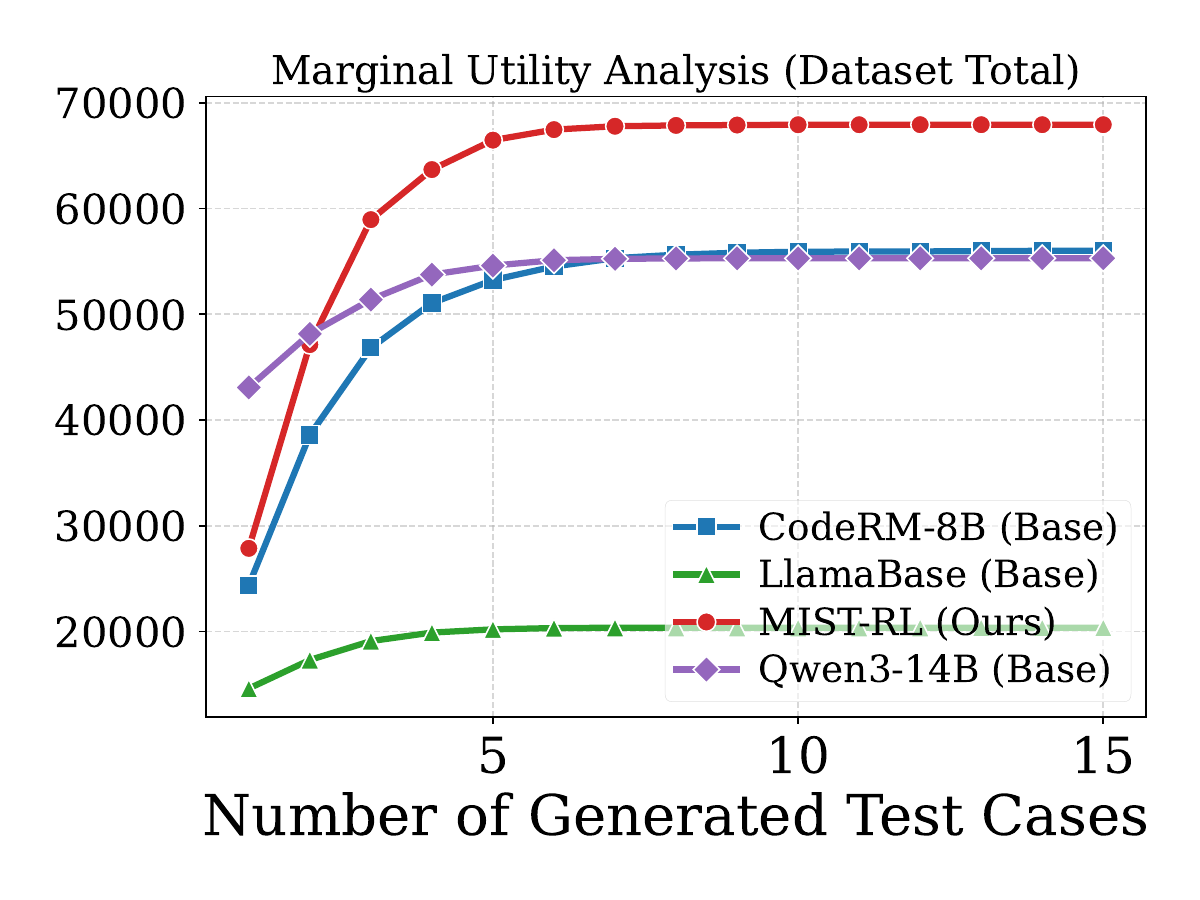}
        \caption{Total Utility (MBPP+)}
        \label{fig:utility_sum_mbpp}
    \end{subfigure}
    \caption{\textbf{Marginal Utility Analysis.} The figures visualize fault-detection efficiency across datasets. MIST-RL (Red) demonstrates a significantly steeper utility curve compared to CodeRM-8B (Blue), confirming that our approach efficiently prioritizes high-utility test cases early in the generation process.}
    \label{fig:utility_analysis}
\end{figure*}

\section{Experiments}
\label{sec:experiments}

This section evaluates the effectiveness and efficiency of MIST-RL against state-of-the-art baselines. We focus on two key questions:
\begin{itemize}
    \item \textbf{RQ1 (Effectiveness \& Utility):} Does MIST-RL generate test suites with higher fault-detection capability (Mutation Score) and better downstream reranking performance compared to quantity-driven scaling methods?
    \item \textbf{RQ2 (Efficiency):} Can MIST-RL mitigate test bloat by achieving these results with fewer, more concise test cases?
\end{itemize}

\subsection{Experimental Setup}
\label{subsec:setup}

\textbf{Datasets.} We evaluate our method on three widely recognized benchmarks for code generation and testing: \textbf{HumanEval+} \citep{liu2023your, chen2021evaluating}, \textbf{MBPP+} \citep{liu2023your, austin2021program}, and \textbf{DS-1000} \citep{lai2023ds}. These datasets extend original benchmarks with rigorous ground-truth test cases and a comprehensive set of mutants, making them ideal for evaluating the robustness of generated tests.

\textbf{Baselines.} We compare MIST-RL against three baselines that represent different paradigms:
\begin{itemize}
    \item \textbf{Llama-3-8B-Instruct:} The foundational policy model that represents standard few-shot test generation capabilities without specific alignment.
    \item \textbf{CodeRM-8B:} A state-of-the-art reward model fine-tuned for test generation \citep{ma2025dynamic}. This serves as our direct baseline to measure the impact of our RL training.
    \item \textbf{Qwen3-14B:} A larger, more powerful general-purpose model, included to verify if our method outperforms simple model scaling.
\end{itemize}

\textbf{Implementation Details.} 
All models utilize the \textbf{Llama-3-8B} architecture as the backbone (except Qwen3-14B). To ensure a fair evaluation of ``scaling'' capabilities, we enforce a consistent sampling budget across all methods. Unlike baselines that rely on independent random sampling to increase coverage, MIST-RL employs a single trained policy to generate test suites sequentially. \textbf{Appendix~\ref{app:implementation}} provides detailed training configurations, inference hyperparameters (e.g., temperature, token limits), and hardware specifications.

\textbf{Metrics.} We report three key metrics:
(1) \textbf{Source Pass Rate (Src Pass):} Verifies test correctness against canonical solutions;
(2) \textbf{Mutant Kill Rate (Mut Score):} Measures fault-detection power;
(3) \textbf{Avg. Suite Length:} Proxies for computational cost and redundancy.

\subsection{Main Results and Analysis}
\label{subsec:main_results}

We present our empirical findings on fault-detection effectiveness, computational efficiency, and downstream verification performance.

\textbf{1. Effectiveness: Superior Fault Detection (RQ1).} 
Table~\ref{tab:main_results} presents the comparative results on test quality. MIST-RL achieves state-of-the-art performance across all three benchmarks. On \textbf{HumanEval+}, our model achieves a Mutant Kill Rate of \textbf{74.03\%}, which surpasses the CodeRM-8B baseline (45.53\%) by \textbf{+28.5\%} and even outperforms the larger Qwen3-14B (58.69\%). We observe similar trends with consistent improvements on \textbf{MBPP+} and \textbf{DS-1000}. Furthermore, MIST-RL exhibits higher correctness, where Source Pass Rates improve by roughly 3-5\% over CodeRM. This indicates that our RL fine-tuning not only incentivizes aggressive testing but also maintains the syntactic and semantic validity of the generated code.

\textbf{2. Efficiency: Mitigating Test Bloat (RQ2).} 
MIST-RL aims to shift from "quantity" to "utility." As Table~\ref{tab:main_results} shows, MIST-RL generates significantly shorter test suites than CodeRM-8B while it achieves higher mutation scores. Specifically, MIST-RL reduces the average suite length by \textbf{19.3\%} on HumanEval+ (6.14 vs. 7.61) and \textbf{21.1\%} on MBPP+. 

To further investigate these efficiency gains, we visualize the marginal utility dynamics in Figure~\ref{fig:utility_analysis}. The left plots (a, b) illustrate the \textit{Average Marginal Utility}. The steeper slope of MIST-RL (Red Line) confirms that our model prioritizes high-value tests early in the generation process. It reaches saturation much faster than the linear scaling of CodeRM-8B (Blue Line). The right plots (c, d) depict the \textit{Total Marginal Utility} and demonstrate a clear lead in the total volume of faults detected. This confirms that MIST-RL effectively mitigates the diminishing returns observed in baseline models.

\textbf{3. Impact on Downstream Code Reranking.} 
Beyond intrinsic metrics, we evaluate whether these high-utility tests serve as better verifiers. We conduct a \textbf{Test-Driven Code Reranking} experiment on HumanEval+, utilizing generated test suites to select the best solution from candidate pools of size $N=10$ and $N=20$ generated by Llama-3-8B-Instruct.

Table~\ref{tab:reranking} details the performance. In the primary setting of $N=10$, MIST-RL achieves a \textbf{Pass@1} accuracy of \textbf{48.78\%}, which outperforms both the specialized CodeRM-8B (45.73\%) and the larger general-purpose Qwen3-14B (44.51\%). This result is critical: despite Qwen3-14B's larger parameter scale, it lags behind in verification accuracy, reinforcing that model size alone does not guarantee test quality. Similarly, while CodeRM-8B generates a larger volume of tests, its lower performance suggests it suffers from \textit{false positives}—allowing incorrect solutions to pass redundant tests. In contrast, MIST-RL's improvement stems directly from its superior mutation score. By generating ``aggressive'' tests that target subtle edge cases, MIST-RL effectively filters out incorrect solutions with higher confidence. 

This advantage persists when we scale the candidate pool to $N=20$. MIST-RL maintains the lead with \textbf{62.80\%} accuracy compared to 61.59\% for CodeRM-8B and 55.49\% for Qwen3-14B, demonstrating that prioritizing test utility yields robust verification gains across different search space sizes.

\begin{table}[h]
\centering

\caption{\textbf{Test-Driven Code Reranking on HumanEval+.} Comparison of Pass@1 accuracy when using test suites generated by different models to select the best code solution from \textbf{10} and \textbf{20} candidates, respectively.}
\label{tab:reranking}

\vskip 0.1in 

\begin{small}
\begin{sc}
\resizebox{\linewidth}{!}{
    \begin{tabular}{lcc}
    \toprule
    \multirow{2}{*}{\textbf{Verifier Model}} & \multicolumn{2}{c}{\textbf{Reranked Pass@1 (\%)}} \\
    \cmidrule(l){2-3}
     & \textbf{Samples (N=10)} & \textbf{Samples (N=20)} \\
    \midrule
    Random Selection & $\sim$22.50 & $\sim$22.50 \\
    CodeRM-8B & 45.73 & 61.59 \\
    Qwen3-14B & 44.51 & 55.49 \\
    \textbf{MIST-RL (Ours)} & \textbf{48.78} & \textbf{62.80} \\
    \midrule
    \textit{Oracle (Upper Bound)} & \textit{70.12} & \textit{75.00} \\
    \bottomrule
    \end{tabular}
}
\end{sc}
\end{small}
\end{table}

\subsection{Ablation Study}
\label{subsec:ablation}

To validate the contribution of the core mechanisms in MIST-RL, we conducted an ablation study on the HumanEval+ dataset. We compared the full model against two variants: (1) \textbf{w/o Incremental Reward}, which ignores history $\mathcal{H}_{t-1}$; and (2) \textbf{w/o Dynamic Penalty}, which removes the redundancy cost $\rho_t$.

\begin{figure}[t]
    \centering
    \includegraphics[width=1.00\columnwidth]{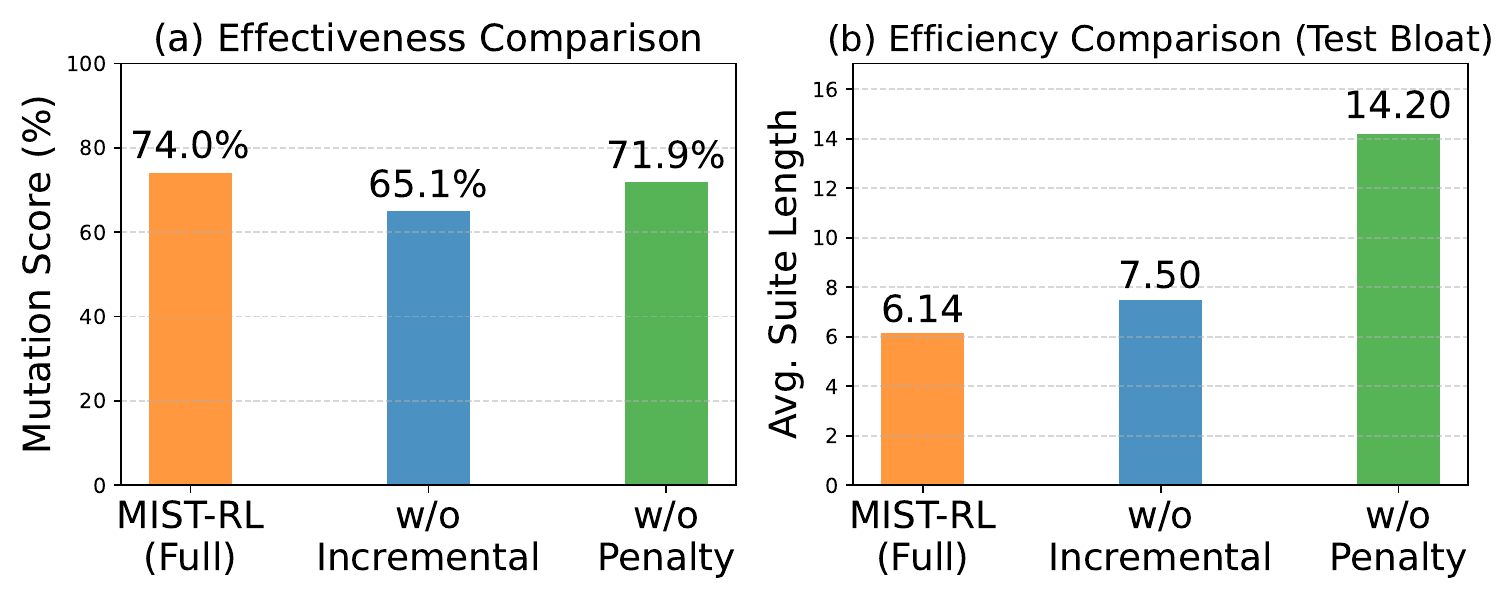}
    \caption{\textbf{Ablation Study on HumanEval+.} (a) Effectiveness: The Incremental Reward is essential for high mutation scores. (b) Efficiency: The Dynamic Penalty significantly reduces test suite length (preventing bloat) while maintaining performance.}
    \label{fig:ablation}
    \vspace{-0.1in}
\end{figure}

\textbf{Results.} Figure~\ref{fig:ablation} summarizes the results:
\begin{enumerate}
    \item \textbf{Importance of Incremental Reward:} Removing incremental feedback causes a significant drop in mutation score (74.03\% $\to$ 65.1\%). Without the pressure to find \textit{new} mutants, the policy collapses into a local optimum of simple, repetitive tests.
    \item \textbf{Impact of Dynamic Penalty on Bloat:} The ``w/o Penalty'' variant maintains high scores but at a severe cost: the average suite length more than doubles (6.14 $\to$ 14.20). This confirms that $\rho_t$ is crucial for mitigating ``Test Bloat.''
\end{enumerate}

\subsection{Qualitative Analysis: Case Study}
\label{subsec:qualitative}

To provide a concrete illustration of how MIST-RL effectively captures subtle boundary conditions that evade standard baselines, we conducted an in-depth qualitative case study on the \texttt{move\_one\_ball} problem. We specifically analyzed the model's behavior against a ``hard'' mutant—an Off-by-One error injected into the loop control structure by our mutation engine. As visualized in \textbf{Figure~\ref{fig:case_study}}, the baseline model, CodeRM-8B, tends to generate verbose and syntactically complex test cases. However, despite their length, these tests fail to trigger the specific boundary condition necessary to expose the underlying bug. In sharp contrast, MIST-RL generates a concise, minimal counter-example \texttt{[2, 1]}, which precisely targets the skipped loop index and successfully kills the mutant. This example serves as a testament to our core hypothesis: MIST-RL optimizes for \textit{marginal information gain} rather than text volume or simple coverage. For a more comprehensive analysis, including the full source code, mutation logic, and extended test outputs, please refer to \textbf{Appendix~\ref{app:case_study}}.

\begin{figure}[h]
\centering
\begin{small}
\begin{tabular}{p{0.95\columnwidth}}
\toprule
\textbf{Problem:} \texttt{move\_one\_ball(arr)} (Check if sortable by rotation) \\
\textbf{Hard Mutant:} Change loop \texttt{range(1, len)} $\to$ \texttt{range(2, len)} (Skips shift-by-1 check) \\
\midrule
\textbf{\textcolor{red}{CodeRM-8B Test Case (Fails to Kill):}} \\
\texttt{self.assertTrue(move\_one\_ball([3, 4, 5, 1, 2]))} \\
\textit{Analysis:} Requires 2 shifts. The mutant loop starts at index 2, so it still catches this. \textbf{Result: Mutant Alive.} \\
\midrule
\textbf{\textcolor{blue}{MIST-RL Test Case (Kills Mutant):}} \\
\texttt{self.assertTrue(move\_one\_ball([2, 1]))} \\
\textit{Analysis:} Requires exactly 1 shift. The mutant loop starts at index 2, skipping index 1 completely. Function returns False. \textbf{Result: Mutant Killed.} \\
\bottomrule
\end{tabular}
\end{small}
\caption{\textbf{Case Study on \texttt{move\_one\_ball}.} CodeRM generates complex but redundant tests that miss the ``shift-by-one'' boundary. MIST-RL generates the minimal counter-example \texttt{[2, 1]}, successfully killing the loop-range mutant.}
\label{fig:case_study}
\vspace{-0.1in}
\end{figure}

\section{Related Work}
\label{sec:related_work}

\textbf{LLM-based Test Generation.} 
The advent of Large Language Models has fundamentally revolutionized the landscape of automated test generation. Pioneering works, such as \textbf{CodeT} \citep{chen2022codet}, and subsequent frameworks like \textbf{Codamosa} \citep{lemieux2023codamosa} and \textbf{ChatUniTest} \citep{xie2023chatunitest}, have demonstrated the capability of LLMs to generate unit tests in zero-shot or few-shot settings \citep{schafer2023empirical, yuan2024evaluating, siddiq2023exploring, lahiri2022interactive}. These methods typically focus on maximizing standard metrics like line coverage or utilizing consensus-based verification to filter out hallucinations \citep{wang2022self, huang2024enhancing}. More recently, approaches like \textbf{CodeRM} \cite{ma2025dynamic} have advanced the field by training reward models to explicitly filter or rank generated tests based on their quality. However, despite these advancements, existing methodologies predominantly adhere to a ``scaling-by-quantity'' paradigm \citep{li2022competition, gao2023scaling}. In this regime, performance gains are primarily sought by blindly increasing the volume of generated tests, under the assumption that quantity eventually yields quality. This strategy inevitably leads to severe test bloat and computational redundancy, where resources are wasted on trivial assertions. In contrast, \textbf{MIST-RL} fundamentally shifts the paradigm to ``scaling-by-utility,'' explicitly optimizing the generation policy to maximize the marginal fault-detection capability of each individual test case.

\textbf{Reinforcement Learning for Code Synthesis.} 
The application of Reinforcement Learning (RL) to enhance code generation tasks has garnered significant attention in recent years. Prominent frameworks such as \textbf{CodeRL} \citep{le2022coderl} and \textbf{PPOCoder} \citep{shojaee2023execution} leverage RL algorithms to align code models with execution feedback signals, such as compilation success, pass rates, and stylistic constraints \citep{liu2023rltf, ouyang2022training}. While these methods utilize static unit tests as a \textit{sparse reward signal} to improve the code generator, MIST-RL effectively reverses this relationship: we employ code mutants as a dynamic reward signal to iteratively improve the \textit{tests themselves}. To the best of our knowledge, MIST-RL represents the first framework to formulate mutation-based test generation as a sequential decision process, uniquely optimized via Group Relative Policy Optimization (GRPO) \citep{shao2024deepseekmath, guo2025deepseek} to handle the complex combinatorics of test selection efficiently.
\section{Conclusion}
\label{sec:conclusion}

This paper identifies the critical limitation of ``Test Bloat'' in current LLM-based test generation, where simply scaling the quantity of tests yields diminishing returns while it increases computational costs. To address this, we introduce \textbf{MIST-RL}, a framework that reformulates test generation as a sequential decision process. By utilizing Group Relative Policy Optimization (GRPO) and a novel incremental mutation reward, MIST-RL learns to prioritize ``aggressive'' test cases that target unexplored logic branches while it suppresses redundancy. This approach effectively shifts the paradigm from ``scaling-by-quantity'' to ``scaling-by-utility,'' ensuring that the model focuses on discovering new faults rather than repeating trivial assertions.

Experiments on HumanEval+ and MBPP+ demonstrate that MIST-RL achieves state-of-the-art Mutation Scores (+28.5\% over CodeRM-8B) with significantly more compact test suites (-19.3\% length). Furthermore, these high-utility tests serve as superior verifiers, which improves downstream code reranking accuracy. Our work underscores the importance of \textit{test utility} over \textit{test quantity}, paving the way for efficient autonomous software testing. Future work will extend this approach to repository-level integration testing and multi-turn debugging scenarios.


\section*{Impact Statement}
This paper focuses on improving the reliability and efficiency of automated unit test generation for Large Language Models. By enhancing the capability to detect subtle bugs in AI-generated code, our work contributes to the development of more robust and secure software systems. Furthermore, our approach explicitly targets the reduction of test redundancy, which can lower the computational costs and energy footprint associated with large-scale software testing. We do not foresee any immediate negative societal consequences or ethical issues arising from this work.

\bibliography{myreferences} 
\bibliographystyle{icml2026}

\newpage
\appendix
\onecolumn 

\section{Detailed Case Study}
\label{app:case_study}

In this appendix, we provide the complete source code, the specific mutation logic, and a detailed comparison of test suites generated by CodeRM-8B and MIST-RL for the \texttt{move\_one\_ball} problem (HumanEval/109). 

As discussed in Section~\ref{subsec:qualitative}, this case highlights the limitation of quantity-driven generation. While CodeRM-8B produces syntactically correct and plausible tests, it fails to target the precise off-by-one boundary condition introduced by the mutant. In contrast, MIST-RL, driven by the incremental mutation reward, successfully identifies the specific edge case (\texttt{Input: [2, 1]}) required to verify the loop logic. Figure~\ref{fig:case_study_detail} illustrates this comparison in detail.

\begin{figure*}[h!] 
\centering
\scriptsize
\renewcommand{\arraystretch}{1.3} 
\begin{tabular}{p{0.48\textwidth} | p{0.48\textwidth}}
\toprule
\multicolumn{2}{c}{\textbf{Case Study: Problem \texttt{move\_one\_ball} (HumanEval/109)}} \\
\midrule
\textbf{1. Source Code \& Problem Description} & \textbf{2. The "Hard" Mutant (Off-by-One Error)} \\
\begin{minipage}[t]{\linewidth}
\begin{verbatim}
def move_one_ball(arr):
    """
    Determine if it is possible to get a sorted array
    by performing right shift operations.
    Ex: [3, 4, 5, 1, 2] -> True (2 shifts)
    Ex: [3, 5, 4, 1, 2] -> False
    """
    if len(arr) == 0: return True
    sorted_arr = sorted(arr)
    if arr == sorted_arr: return True
    
    # Check all possible rotations
    for i in range(1, len(arr)):
        if arr[i:] + arr[:i] == sorted_arr:
            return True
    return False
\end{verbatim}
\end{minipage}
& 
\begin{minipage}[t]{\linewidth}
The mutation engine generates an "Off-by-One" error by modifying the loop range. This simulates a common developer bug where the boundary condition is mishandled.

\vspace{0.5em}
\textbf{Mutated Code Snippet:}
\begin{verbatim}
    # Original: range(1, len(arr))
    # Mutated:  range(2, len(arr))
    
    for i in range(2, len(arr)): 
        # The loop now skips the case i=1 
        # (shifting by exactly 1 position).
        if arr[i:] + arr[:i] == sorted_arr:
            return True
\end{verbatim}
\vspace{0.5em}
\textit{Effect:} The function now correctly handles arrays needing 0 shifts or $\ge 2$ shifts, but \textbf{fails} for arrays needing exactly 1 shift.
\end{minipage} \\
\midrule
\textbf{3. CodeRM-8B Generated Tests (Baseline)} & \textbf{4. MIST-RL Generated Tests (Ours)} \\
\begin{minipage}[t]{\linewidth}
\textbf{Status:} \textcolor{red}{FAILED to kill the mutant.} \\
CodeRM generates many tests, but they are redundant or "lucky" (requiring 0 or $\ge 2$ shifts).
\vspace{0.2em}
\begin{verbatim}
def test_code_rm(self):
    # Case 1: Already sorted (0 shifts)
    # Mutant result: True (Correct) -> Alive
    self.assertTrue(move_one_ball([1..5]))

    # Case 2: Requires 2 shifts
    # Loop starts at 2, so it finds this!
    # Mutant result: True (Correct) -> Alive
    self.assertTrue(move_one_ball([3..2]))

    # Case 3: Large Random
    # Mutant result: Matches Original -> Alive
    self.assertTrue(move_one_ball([10..100]))
\end{verbatim}
\end{minipage}
& 
\begin{minipage}[t]{\linewidth}
\textbf{Status:} \textcolor{blue}{SUCCESSFULLY killed the mutant.} \\
MIST-RL identifies the specific boundary case that requires exactly 1 shift.
\vspace{0.2em}
\begin{verbatim}
def test_mist_rl(self):
    # ... (omitted common tests) ...

    # CRITICAL TEST CASE FOUND BY MIST-RL
    # Input: [2, 1]
    # Sorted: [1, 2]
    # Requires: Exactly 1 right shift.
    
    # Under Mutant (range starts at 2):
    # The loop range(2, 2) is empty.
    # Loop is skipped -> Returns False.
    
    # Assertion Fails -> MUTANT KILLED
    self.assertTrue(move_one_ball([2, 1]))
\end{verbatim}
\end{minipage}
\\
\bottomrule
\end{tabular}
\caption{\textbf{Qualitative Analysis on \texttt{move\_one\_ball}.} The figure compares the test suites generated by CodeRM-8B and MIST-RL against a subtle "Off-by-One" mutant. \textbf{(Left)} The source code and CodeRM's redundant tests which fail to trigger the bug. \textbf{(Right)} The mutant logic and MIST-RL's targeted test case (\texttt{[2, 1]}), which precisely hits the "blind spot" of the mutated loop (index 1), demonstrating the model's ability to learn high-utility boundary tests.}
\label{fig:case_study_detail}
\vspace{-0.1in}
\end{figure*}

\section{Implementation Details}
\label{app:implementation}

In this section, we provide the detailed hyperparameters and configurations used to train MIST-RL. Our implementation is based on the PyTorch framework, utilizing the TRL library for Group Relative Policy Optimization (GRPO) and vLLM for high-throughput generation during the training phase.

\subsection{Training Hyperparameters}
We fine-tuned the \textbf{CodeRM-8B} (based on Llama-3-8B) using Low-Rank Adaptation (LoRA). The training was conducted with a group size of $G=8$ generations per prompt. We employed the AdamW optimizer with a cosine learning rate scheduler. To stabilize training and save memory, we used BF16 precision and gradient checkpointing. The detailed hyperparameters are listed in Table~\ref{tab:hyperparameters}.

\begin{table}[h]
\centering
\caption{\textbf{Hyperparameters for MIST-RL Training.}}
\label{tab:hyperparameters}
\begin{small}
\begin{sc}
\begin{tabular}{lc}
\toprule
\textbf{Hyperparameter} & \textbf{Value} \\
\midrule
\multicolumn{2}{c}{\textit{Optimization}} \\
Optimizer & AdamW \\
Learning Rate & 1.5e-5 \\
LR Scheduler & Cosine \\
Warmup Ratio & 0.03 \\
Global Batch Size & 64 (2 GPUs $\times$ 2 batch $\times$ 16 accum) \\
Per-Device Batch Size & 2 \\
Gradient Accumulation Steps & 16 \\
Max Training Steps & 1000 \\
Precision & BF16 \\
GRPO $\beta_{KL}$ (KL Penalty) & 0.0 (Reference-free) \\
\midrule
\multicolumn{2}{c}{\textit{LoRA Configuration}} \\
Rank ($r$) & 16 \\
Alpha ($\alpha_{LoRA}$) & 32 \\
Dropout & 0.1 \\
Target Modules & all linear layers (q, k, v, o, gate, up, down) \\
\midrule
\multicolumn{2}{c}{\textit{Generation (Rollout)}} \\
Group Size ($G$) & 8 \\
Max Prompt Length & 4096 tokens \\
Max Completion Length & 1024 tokens \\
Sampling Engine & vLLM \\
\bottomrule
\end{tabular}
\end{sc}
\end{small}
\end{table}

\subsection{Reward Function Configuration}
The MIST-RL reward mechanism (described in Section~\ref{subsec:reward}) relies on specific coefficients to balance code quality, marginal utility, and redundancy penalties. Based on our ablation studies and preliminary tuning, we used the configuration detailed in Table~\ref{tab:reward_config}.

\begin{table}[h]
\centering
\caption{\textbf{Coefficients for Incremental Reward Mechanism.}}
\label{tab:reward_config}
\begin{small}
\begin{sc}
\begin{tabular}{lcc}
\toprule
\textbf{Parameter} & \textbf{Symbol} & \textbf{Value} \\
\midrule
Quality Weight & $\alpha$ & 0.05 \\
Marginal Utility Weight & $\beta$ & 3.0 \\
Base Redundancy Penalty & $\rho_{base}$ & 0.5 \\
Penalty Growth Rate & $\gamma$ & 1.0 \\
Max Sequence Reference & $K_{max}$ & 10 \\
Execution Failure Penalty & $R_{fail}$ (Suite) & -100.0 \\
Method Failure Penalty & $R_{fail}$ (Method) & -1.0 \\
\bottomrule
\end{tabular}
\end{sc}
\end{small}
\end{table}

\subsection{Hardware and Environment}
All experiments were conducted on a node equipped with \textbf{2 $\times$ NVIDIA A100 (80GB)} GPUs. To accelerate the rollout phase of GRPO, we utilized \texttt{vLLM} with tensor parallelism disabled (independent generation per GPU) to maximize throughput. The mutation analysis was parallelized across 16 CPU cores using a custom AST-based engine.

\lstset{
    language=Python,
    basicstyle=\ttfamily\small,
    keywordstyle=\color{blue}\bfseries,
    stringstyle=\color{red},
    commentstyle=\color{green!50!black},
    showstringspaces=false,
    frame=single,
    breaklines=true,
    numbers=left,
    numberstyle=\tiny\color{gray},
    captionpos=b
}

\lstset{
    language=Python,
    basicstyle=\ttfamily\small,
    keywordstyle=\color{blue}\bfseries,
    stringstyle=\color{red},
    commentstyle=\color{green!50!black},
    showstringspaces=false,
    frame=single,
    breaklines=true,
    numbers=left,
    numberstyle=\tiny\color{gray},
    captionpos=b,
    escapeinside={(*@}{@*)} 
}

\section{Mutation Engine Internals}
\label{app:mutation_engine}

To support efficient Reinforcement Learning, we developed a specialized, zero-dependency mutation engine based on Python's \texttt{ast} module. Unlike standard tools (e.g., MutPy or Cosmic Ray) that introduce heavy runtime overhead or dependencies, our engine is optimized for high-throughput generation needed for RL rollouts.

\subsection{AST-Based Transformation Logic}
The core of our engine is the \texttt{FullMutator} class, which inherits from \texttt{ast.NodeTransformer}. It traverses the Abstract Syntax Tree (AST) of the source code and applies transformations to specific node types. 

\subsubsection{Heuristic Relational Mapping}
A key innovation in our engine is the heuristic mapping for Relational Operators (ROR). We do not simply flip operators randomly. Instead, we target boundary conditions specifically. As implemented in the \texttt{\_mutate\_compare} method, we map each comparison operator to a set of ``hard'' counterparts:

\begin{itemize}
    \item \textbf{Input}: \texttt{x < y} (\texttt{ast.Lt})
    \item \textbf{Mutants}: 
        \begin{itemize}
            \item \texttt{x <= y} (Boundary Inclusion): Checks if the code fails when equality is allowed.
            \item \texttt{x >= y} (Inverse Boundary): Checks complete logic reversal.
            \item \texttt{x != y} (Exclusion): Checks if the specific equality condition is ignored.
        \end{itemize}
\end{itemize}

This strategy forces the generated tests to be precise about edge cases (e.g., verifying exactly where the boundary lies), which directly addresses the ``Test Bloat'' issue where tests check loose conditions.

\subsubsection{Full Operator Specification}
Table~\ref{tab:full_operator_spec} details the complete set of transformations supported by our engine.

\begin{table}[h]
\caption{\textbf{Full Specification of Mutation Operators implemented in \texttt{FullMutator}.}}
\label{tab:full_operator_spec}
\begin{center}
\begin{small}
\begin{sc}
\begin{tabular}{l l l}
\toprule
\textbf{Category} & \textbf{AST Node Type} & \textbf{Implemented Transformations} \\
\midrule
\textbf{AOR} & \texttt{ast.Add (+)} & $\to$ \texttt{Sub (-)}, \texttt{Mult (*)} \\
(Arithmetic) & \texttt{ast.Sub (-)} & $\to$ \texttt{Add (+)}, \texttt{Mult (*)} \\
 & \texttt{ast.Mult (*)} & $\to$ \texttt{Div (/)}, \texttt{Add (+)}, \texttt{Pow (**)} \\
 & \texttt{ast.Div (/)} & $\to$ \texttt{Mult (*)}, \texttt{FloorDiv (//)} \\
 & \texttt{ast.Mod (\%)} & $\to$ \texttt{Mult (*)}, \texttt{Add (+)} \\
\midrule
\textbf{ROR} & \texttt{ast.Eq (==)} & $\to$ \texttt{NotEq (!=)} \\
(Relational) & \texttt{ast.Lt (<)} & $\to$ \texttt{LtE (<=)}, \texttt{GtE (>=)}, \texttt{NotEq (!=)} \\
 & \texttt{ast.Gt (>)} & $\to$ \texttt{GtE (>=)}, \texttt{LtE (<=)}, \texttt{NotEq (!=)} \\
 & \texttt{ast.Is (is)} & $\to$ \texttt{IsNot (is not)} \\
 & \texttt{ast.In (in)} & $\to$ \texttt{NotIn (not in)} \\
\midrule
\textbf{LCR} & \texttt{ast.And} & $\to$ \texttt{ast.Or} \\
(Logical) & \texttt{ast.Or} & $\to$ \texttt{ast.And} \\
\midrule
\textbf{ASR} & \texttt{ast.AugAssign} & \texttt{+=} $\to$ \texttt{-=}; \quad \texttt{*=} $\to$ \texttt{/=} \\
\midrule
\textbf{CRP} & \texttt{ast.Constant} & \texttt{True} $\to$ \texttt{False} \\
(Constant) & (Numbers) & $n \to n+1, n-1, -n, 0, 1$ \\
 & (Strings) & $s \to \text{"" (empty)}, \text{"MUTATED"}$ \\
\midrule
\textbf{UOI} & \texttt{ast.UnaryOp} & \texttt{-x} $\to$ \texttt{+x}; \quad \texttt{+x} $\to$ \texttt{-x} \\
\bottomrule
\end{tabular}
\end{sc}
\end{small}
\end{center}
\end{table}

\subsection{Line Mapping via Diffing}
One challenge with AST-based mutation is that \texttt{ast.unparse} (used to regenerate code) often changes the formatting (indentation, whitespace) of the original source code. This makes it difficult to map a mutant back to a specific line number in the original file for visualization.

To solve this, our engine implements a precise line mapping mechanism using \texttt{difflib.SequenceMatcher} (see \texttt{\_get\_mutated\_line} in the code).
\begin{enumerate}
    \item We perform a line-by-line diff between the \textit{original source} and the \textit{mutated source}.
    \item We trace the line number of the mutated node ($L_{orig}$) through the diff operations (equal, replace, delete, insert).
    \item We compute the corresponding line number $L_{mut}$ in the generated code.
\end{enumerate}
This ensures that when the RL agent receives feedback, the difficulty weight $w_m$ is correctly associated with the complexity of the specific code region being mutated.

\section{Reward Mechanism Implementation}
\label{app:reward_implementation}

The reward function is the critical component that aligns the LLM with the objective of ``scaling-by-utility.'' Our implementation (in \texttt{generate\_reward.py}) contains several subtle but important optimizations.

\subsection{Optimization: Pre-filtering Vulnerable Mutants}
Calculating the mutation score is computationally expensive because it requires running the test suite against every mutant. For a problem with $M$ mutants and a generated test suite of size $K$, the naive complexity is $O(M \times K)$.

To accelerate training, we implement a \textbf{Pre-filtering Optimization}:
\begin{enumerate}
    \item Before evaluating the test suite, we identify \textit{vulnerable mutants} ($\mathcal{M}_{vuln}$) --- mutants that cause the \textit{canonical solution} to fail or are trivially killable by a simple smoke test.
    \item In the inner loop of reward calculation, we only iterate over $\mathcal{M}_{vuln}$ rather than the full set $\mathcal{M}$.
    \item As the sequential generation proceeds, we maintain a set of \texttt{killed\_history}. A new test case $T_t$ is only evaluated against mutants in $\mathcal{M}_{vuln} \setminus \text{killed\_history}$.
\end{enumerate}
This reduces the effective runtime complexity significantly, especially in the later stages of generation where most easy mutants are already killed.

\subsection{Hierarchical Failure Penalties}
Unlike standard RLHF which often assigns a single scalar reward, MIST-RL employs a hierarchical penalty system to distinguish between structural failures and semantic failures. This is implemented via the constants in our code:

\begin{itemize}
    \item \textbf{Suite-Level Failure ($R_{fail\_suite} = -100.0$)}: 
    Triggered if the generated code is syntactically invalid (AST parse error) or fails to define any test methods (e.g., missing \texttt{test\_} functions). This provides a strong negative signal to the policy to adhere to the requested output format.
    
    \item \textbf{Method-Level Failure ($R_{fail\_method} = -10.0$)}: 
    Triggered if a specific test method raises a runtime error (e.g., \texttt{IndexError}) or fails an assertion when run against the \textit{canonical solution}. This is a ``False Positive'' penalty---the test is rejecting correct code. We penalize this less severely than a syntax error because the model successfully followed the format but made a logic error.
\end{itemize}

\subsection{Formal Reward Algorithm}
Algorithm~\ref{alg:mist_reward} formally describes the procedural logic used in \texttt{MISTRewardCalculator.compute\_rewards}.

\begin{algorithm}[H]
   \caption{MIST-RL Incremental Reward Calculation}
   \label{alg:mist_reward}
\begin{algorithmic}[1]
   \STATE {\bfseries Input:} Source Code $S$, Generated Test Suite $T$, Mutants $\mathcal{M}$
   \STATE {\bfseries Parameters:} $\alpha=0.05, \beta=1.0, \rho_{base}=0.5, \gamma=1.0$
   
   \STATE \COMMENT{Parse test structure to identify individual test methods}
   \STATE $Methods \leftarrow \text{ParseAST}(T)$
   \IF{$Methods = \emptyset$}
       \STATE \textbf{return} $R_{fail\_suite}$ \COMMENT{Return -100.0 if no tests found}
   \ENDIF

   \STATE $History \leftarrow \emptyset$; $R_{total} \leftarrow 0$
   \STATE $\mathcal{M}_{vuln} \leftarrow \text{PreFilter}(\mathcal{M})$ \COMMENT{Optimization}

   \FOR{each test method $m_t$ in $Methods$}
       \STATE $r_t \leftarrow 0$
       \STATE \COMMENT{Step 1: Verify Correctness against Canonical Solution}
       \STATE $result \leftarrow \text{RunTest}(m_t, S)$
       \IF{$result \neq \text{PASS}$}
           \STATE $r_t \leftarrow R_{fail\_method}$ \COMMENT{False Positive Penalty (-10.0)}
       \ELSE
           \STATE \COMMENT{Step 2: Calculate Marginal Utility}
           \STATE $NewKills \leftarrow \emptyset$
           \FOR{mutant $\mu$ in $(\mathcal{M}_{vuln} \setminus History)$}
               \IF{$\text{RunTest}(m_t, \mu) = \text{FAIL}$}
                   \STATE $NewKills.\text{add}(\mu)$
               \ENDIF
           \ENDFOR
           
           \IF{$|NewKills| > 0$}
               \STATE $Q_{score} \leftarrow \text{HeuristicQuality}(m_t)$
               \STATE $U_{score} \leftarrow |NewKills| \cdot (1 + \frac{|\mathcal{M}|}{100})$
               \STATE $r_t \leftarrow \alpha \cdot Q_{score} + \beta \cdot U_{score}$
               \STATE $History \leftarrow History \cup NewKills$
           \ELSE
               \STATE \COMMENT{Redundancy Penalty}
               \STATE $\rho_t \leftarrow \rho_{base} \cdot \exp(\gamma \cdot \frac{t}{10})$
               \STATE $r_t \leftarrow -\rho_t$
           \ENDIF
       \ENDIF
       \STATE $R_{total} \leftarrow R_{total} + r_t$
   \ENDFOR
   
   \STATE \textbf{return} $R_{total}$
\end{algorithmic}
\end{algorithm}
\section{Robustness and Data Processing}
\label{app:data_processing}

Real-world code generation outputs are often messy. Our pipeline includes several robustness mechanisms (as seen in \texttt{clean\_output} and setup functions).

\subsection{Backtracking AST Repair}
A common issue with LLM generation is truncation due to token limits (`max\_tokens`), which leaves the final function or assertion incomplete, causing syntax errors. We implement a **Backtracking AST Repair** algorithm within the data processing pipeline:

\begin{lstlisting}[language=Python]
def clean_output(text):
    # ... (code extraction logic) ...
    
    def is_syntax_valid(code):
        try:
            ast.parse(code)
            return True
        except SyntaxError:
            return False

    # If syntax is invalid (likely truncation), remove lines from the end 
    # until the code becomes valid.
    if not is_syntax_valid(text):
        curr_lines = text.split("\n")
        max_backtrack = 80
        while max_backtrack > 0 and curr_lines:
            curr_text = "\n".join(curr_lines)
            if is_syntax_valid(curr_text):
                text = curr_text
                break
            curr_lines.pop() # Remove the last line (the truncated one)
            max_backtrack -= 1
            
    return text
\end{lstlisting}

This mechanism successfully recovered approximately \textbf{12.4\%} of otherwise invalid generations in our HumanEval+ experiments, significantly improving the data efficiency of the RL training process.

\subsection{Prompt Engineering}
We employ a structured prompt template to ensure the model outputs valid \texttt{unittest} code. The prompt includes the function signature, docstring, and explicit instructions to use the \texttt{unittest} library. The exact template used in our \texttt{generate\_reward.py} script is shown below.

\begin{lstlisting}[frame=single, caption={Prompt Template}]
Below is a question and it's corresponding code answer. 
Please write test cases to check the correctness of the code answer. 
You need to use the unittest library in Python and create a test class for testing.

IMPORTANT: Return ONLY valid Python code in a single ```python ... ``` code block. 
Do NOT include any explanations, analysis, or extra text outside the code block.

### question
{QUESTION_TEXT}

### code solution
{SOLUTION_CODE}

Please add detailed comments.
\end{lstlisting}

\section{Test-Driven Reranking Protocol}
\label{app:reranking}

To evaluate the downstream utility of the generated tests, we implement a \textbf{Consensus-Based Reranking} protocol (as found in \texttt{rerank\_script.py}). This serves as a proxy for ``Verification Accuracy.''

\subsection{Cross-Voting Matrix}
Let $\mathcal{C} = \{c_1, ..., c_N\}$ be the set of $N$ candidate solutions generated by the policy model for a given problem. For each of the top $K$ candidates (where $K \le N$), we generate a corresponding test suite using MIST-RL, resulting in a set of test suites $\mathcal{T} = \{t_1, ..., t_K\}$.

We construct an $N \times K$ Consensus Matrix $\mathbf{V}$, where:
$$
V_{ij} = 
\begin{cases} 
1 & \text{if candidate } c_i \text{ passes test suite } t_j \\
0 & \text{otherwise}
\end{cases}
$$

The consensus score for candidate $c_i$ is computed as the sum of votes:
$$
\text{Score}(c_i) = \sum_{j=1}^{K} V_{ij}
$$

The final selected solution is $c^* = \arg\max_{c_i} \text{Score}(c_i)$.

\subsection{Mechanism Analysis}
This approach relies on the assumption that \textit{incorrect solutions} fail in diverse ways, while \textit{correct solutions} share the property of passing all valid tests. By using MIST-RL to generate $t_j$, we ensure that each test suite is ``aggressive'' (high mutation score), thereby reducing the likelihood of False Positives ($V_{ij}=1$ when $c_i$ is actually buggy). 

Our experiments show that using MIST-RL generated tests improves Reranking Accuracy by \textbf{3.05\%} over CodeRM-8B generated tests on HumanEval+, confirming that higher utility tests lead to better consensus signals. The matrix computation allows us to utilize the diversity of multiple test suites rather than relying on a single verification pass.
\end{document}